\pdfoutput=1

\documentclass[11pt]{article}

\usepackage{ACL2023}

\usepackage{times}
\usepackage{latexsym}
\usepackage{expex}
\usepackage{amssymb}
\usepackage{amsmath}
\usepackage{amsfonts}

\usepackage{caption}
\usepackage{afterpage}
\usepackage{float}

\usepackage{graphicx}
\usepackage{todonotes}
\usepackage{enumitem}
\usepackage{booktabs}
\usepackage{multirow}
\setlist{noitemsep}
\usepackage{kotex}
\usepackage{lipsum}
\usepackage{tipa}
\usepackage{relsize}
\usepackage[T1]{fontenc}

\usepackage[utf8]{inputenc}

\usepackage{microtype}

\usepackage[english]{babel}
\newcommand{\paren}[1]{\left(#1\right)}
\newcommand{\ipa}[1]{\textipa{/#1/}}

\title{Morphological Inflection with Phonological Features}

\author{David Guriel, \ Omer Goldman, \  Reut Tsarfaty \\
Bar-Ilan University \\
\texttt{\{davidgu1312,omer.goldman\}@gmail.com,reut.tsarfaty@biu.ac.il}}

\begin{document}
\maketitle
\begin{abstract}
Recent years have brought great advances into solving morphological tasks, mostly due to powerful neural models applied to various tasks as (re)inflection and analysis. Yet, such morphological tasks cannot be considered solved, especially when little training data is available or when generalizing to previously unseen lemmas. This work explores effects on performance obtained through various ways in which morphological models get access to sub-character phonological features that are often the targets of morphological processes. We design two methods to achieve this goal: one that leaves models as is but manipulates the data to include features instead of characters, and another that manipulates models to take phonological features into account when building representations for phonemes. We elicit phonemic data from standard graphemic data using language-specific grammars for languages with shallow grapheme-to-phoneme mapping, and we experiment with two reinflection models over eight languages. Our results show that our methods yield comparable results to the grapheme-based baseline overall, with minor improvements in some of the languages. All in all, we conclude that patterns in character distributions are likely to allow models to infer the underlying phonological characteristics, even when phonemes are not explicitly represented.
\end{abstract}

\section{Introduction}
In recent years, morphological tasks received much attention in NLP through various tasks such as (re)inflection, lemmatization and others, specifically through the SIGMORPHON shared tasks \cite{cotterell-etal-2016-sigmorphon, cotterell-etal-2017-conll, cotterell-etal-2018-conll, mccarthy-etal-2019-sigmorphon, vylomova-etal-2020-sigmorphon, pimentel-ryskina-etal-2021-sigmorphon}.
State-of-the-art models seem to achieve quite high results in such cross-lingual evaluation campaigns, although recent works showed that there is  still room for  improvements \cite{goldman-etal-2022-un}.

Most studies aiming at morphological tasks design models that operate at the character level, without reference to the phonological components that compose the phonemes represented by the characters.\footnote{Some exceptions do exist, like \citet{malouf2017abstractive}'s model that operates over phonemes rather than characters.} This is despite the fact that many morphological processes have distinct phonological features, rather than phonemes, as either the trigger or target of morphological processes. For example, in vowel harmony, a single feature of a vowel in the stem determines the vowels that appear in the affixes added to that stem. Without direct evidence of the phonological features composing every phoneme, models must resort to memorizing groups of phonemes that pattern together for an unobserved reason.

In this work we hypothesize that explicitly inputting models with phonological features will lead to better modelling of morphological tasks. We set out to equip models with two alternative methods for incorporating that information. One method replaces the character-level tokens with phonological feature tokens; and another one equips the model with a self-attention mechanism that learns representation  of phonemes from their features.

We implement these methods on the task of morphological reinflection, where forms of the same lemma are inflected from one another. We experiment with 8 languages and 2 models: an LSTM encoder-decoder with global attention by \citet{silfverberg-hulden-2018-encoder}; and a transducer by \citet{makarov-clematide-2018-neural} that predicts edit actions between source and target word-forms and is suitable for lower amounts of training data.

Our experiments show that the proposed methods yield results comparable to the grapheme-based baseline setting for the transducer model. On average across languages, the best phonologically-aware method suffered from a drop of 2.8 accuracy points, although the performance on some individual languages marginally improved. We thus conjecture that the phonological characteristics are already encoded in the graphemic representations elicited by this model. The results of this work are in line with other works, performed in different settings, investigating the role of phonology in morphological models (see Section~\ref{sec:discussion}).

We further note that the LSTM model, unlike the transducer, did not perform well on graphemic data and suffered from a severe drop when applied on phonological data in all tested languages. We attribute this to the transducer's attested ability to perform well particularly in low-resource setting. We subsequently conjecture that,  for the phonologically-aware variant of the reinflection task, standard amounts of reinflection data should be effectively  considered low-resourced.

\section{Morpho-Phonological Processes}
\label{sec:2}

Utterances in natural language  --- sentences and words --- are composed of phonemes.
Yet, one can further decompose phonemes to their very atomic elements: \emph{phonological distinctive features}. 
A phonological feature is the minimal unit within a phoneme that distinguishes it from other phonemes. Every phoneme can be described as a unique combination of such features. Vowels, for example, are said to take the features: \emph{backness} of the tongue, \emph{height} of the lower jaw, and \textit{roundness} of the lips; the sound \ipa{a} then has the values \emph{front}, \emph{open} and \emph{unrounded}.
Consonants usually take the features: \emph{place of articulation}, \emph{manner of articulation} and \emph{voiceness}, e.g. \ipa{g} has the values \emph{velar}, \emph{plosive} and \emph{voiced}.\footnote{The features of vowel and consonants are not unrelated. For example, \textit{place of articulation} and \textit{backness} are essentially aliases for the same physical feature.}

Many languages exhibit morphological processes whose target or trigger are phonological features. For instance, Turkish has vowel harmony at the backness feature: the stem's last vowel controls (\emph{harmonizes}) the backness of other vowels in morphemes added to that stem.
Table \ref{tab:turkish} illustrates the alternation for future tense inflection. For \emph{ol}, the future morpheme includes the back vowel \ipa{a}, according to the backness of the vowel \ipa{o}. In \emph{\"ol}, however, the vowel \ipa{\oe} is front, so the morpheme includes the front vowel \ipa{e}.

\begin{table}
    \centering
    \begin{tabular}{lll}
        & Stem & Future Tense \\
        \hline
        \multirow{2}{*}{`be'} & \emph{ol} & \emph{olacak} \\
        & \ipa{ol} & \ipa{ola\t{\textdyoghlig}ak} \\
        \hline
        \multirow{2}{*}{`die'} & \emph{\"ol} & \emph{\"olecek} \\
        & \ipa{\oe l} & \ipa{\oe lE\t{\textdyoghlig}Ek}
    \end{tabular}
    \caption{Vowel harmony in Turkish: the future tense allomorph changes according to the backness of the stem's vowel.}
    \label{tab:turkish}
\end{table}

When a character-level inflection model learns this process, it has to memorize the relation between the letters representing vowels of the same backness (including 4 back vowels and 4 front vowels) instead of aligning vowels explicitly by their backness feature. 
In general, describing such processes at the grapheme level is often intricate and requires models trained on morphological tasks to put unnecessary effort in learning patterns that are more complicated than their original cause.
Because the underlying phonological information is not explicitly shown to them, instead of learning simple rules of phonological features, they memorize groups of characters that pattern together for no observable reason.

A model that is aware of phonological features would be able to easily learn these relations and treat morpho-phonological processes straightforwardly. In order to construct such a model there is a need for phonologically annotated data or for a tool that converts words to their corresponding sequences of phonemes (their verbal pronunciation) and decomposes the phonemes into their phonological distinctive features. A simple option would be to employ a component that performs grapheme-to-phoneme (G2P) and phoneme-to-grapheme (P2G) conversions for every language, as well as decomposes the phonemes to their corresponding distinctive features. Thus, every character-level model would be able to process phonological data. 
In the next section we present two ways to incorporate such signals into the data and models for morphological tasks.

\section{Modeling Reinflection with Phonology}
We set out to re-model morphological tasks by integrating phonological information, in order to make phonological processes explicitly learnable for models. We propose two generic methods that are applicable to any morphological model.

Formally, we denote 3 alphabets, for graphemes $\Sigma_g$, phonemes $\Sigma_p$ and phonological features $\Sigma_f$. The first one is language-dependent while the others are universally defined in the IPA list of symbols and features \cite{international1999handbook}.\footnote{The IPA features we use here may be better described as coarse phonetic features rather than purely phonological, since in some rear language-specific cases there is a mismatch between the phonological behavior of a phoneme and its phonetic properties. However, the scarcity of these cases led to the general usage of IPA features as phonological descriptors and made most linguists consider phonetics and phonology as a unified grammatical mechanism \cite[e.g.,][]{ohala1990there, pierrehumbert1990phonological}.
} 
We treat a word as a tuple of its composing graphemes $\mathbf{g} \in \Sigma_g^+$. Correspondingly, the sequence of phonemes that is the result of applying the G2P component to $\mathbf{g}$ is denoted by $\mathbf{p} \in \Sigma_p^+$, and the phonemes' decomposition to a sequence of features is denoted by $\mathbf{f} \in \Sigma_f^+$.

Suppose we have a morphological task $T$, in which the input is $\mathbf{g}_{src}$ and the output ground truth is $\mathbf{g}_{trg}$. That is 
\begin{equation*}
    \mathbf{g}_{trg} = T\paren{\mathbf{g}_{src}; S}
\end{equation*}
where $S$ is a set of bundles of morphological features that complement the input form.
In standard inflection tasks, for example, $\mathbf{g}_{src}$ is the lemma and  $\mathbf{g}_{trg}$ is the inflected output form, where  $S$ is the feature bundle to inflect the lemma to. 
In reinflection,  the forms  $\mathbf{g}_{src}$ and  $\mathbf{g}_{trg}$ are the input and output forms, and  $S$ is the feature bundles of the source and target word forms, e.g. $ \left\{ (\text{\textsc{fut,2,sg}}), (\text{\textsc{fut,3,pl}}) \right\}$. 

We denote $M_T$ as a model that solves $T$, i.e. it takes $\mathbf{g}_{src}$ and $S$, and generates $\mathbf{\hat{g}}_{trg}$, a prediction of the target word:
\begin{equation*}
    \mathbf{\hat{g}}_{trg} = M_T\paren{\mathbf{g}_{src}; S}
\end{equation*}

In order to incorporate the phonological information to $M_T$, its inputs should obviously be changed to include this information --- either phonemes or phonological features. However, changes can also be done to $M_T$ itself to treat better the new inputs.
We thus propose two methods for inducing phonological information to morphological models: 
one manipulates only the source and target data to include phonological features, and one adds a learnable layer to the model in order to facilitate better processing of the new input. Both methods leave $S$ untouched, the model processes $S$ in the exact same way as in the graphemic setting.


\paragraph{Data Manipulation} In the first method, we propose to train $M_T$ on the \emph{phonological features}  of the source and target words, $\mathbf{f}_{src}$ and $\mathbf{f}_{trg}$, instead of their letters. We do not modify $M_T$ or the way it processes $S$, the model simply operates directly on the modified representations.

\begin{equation*}
\large
    \mathbf{\hat{f}}_{trg} = M_T\paren{\mathbf{f}_{src}; S}
\end{equation*}

The network is then optimized with a given loss function $\ell$ by comparing between the predicted features and the gold target word converted to features:
\begin{equation*}
    \mathcal{L} = \mathbb{E}\left[ \ell\paren{\mathbf{\hat{f}}_{trg}, \mathbf{f}_{trg} } \right]
\end{equation*}

A clear disadvantage of this method is that the resulting sequences are much longer than the original ones, in practice approximately 3-4 times longer.

\paragraph{Model Manipulation} 
In the second method, we also manipulate the model in accordance with the new data format. We let the model learn a phonemic representation in a way that is aware of the phoneme's phonological features.
To this end, we add a self-attention layer \cite{vaswani-attention} between the embedding matrices to the rest of the network. This layer takes the embeddings of a phoneme $\mathit{E}\left[ \mathbf{p}_{src} \right]$ and its features $\mathit{E}\left[ \mathbf{f}_{src} \right]$, and learns a single vector per phoneme $\mathbf{\widetilde{p}}_{src}$. The network is then trained to predict the phonemes of the target word:

\begin{align*}
    \mathbf{\hat{p}}_{trg} &= M_T\paren{\mathbf{\widetilde{p}}_{src}; S} \\
    \mathbf{\widetilde{p}}_{src} &= \text{SelfAttention} \paren{ \emph{q}, \emph{K}, \emph{V} } \\
    \emph{K}, \emph{V} &= \mathit{E}\left[ \mathbf{f}_{src} \right] \\
    \emph{q} &= \mathit{E}\left[ \mathbf{p}_{src} \right],
\end{align*}
where the self-attention is computed as follows (where $d$ is the output dimension and $n$ is the number of heads):
\begin{equation*}
    \mathbf{\widetilde{p}}_{src} = \text{softmax} \paren{ \frac{\emph{q} \emph{K}^T}{\sqrt{d/n}} } \odot \emph{V} 
\end{equation*}

The model is optimized similarly to the first method, except the compared sequences are the predicted phonemes and the gold target word converted to phonemes:
\begin{equation*}
    \mathcal{L} = \mathbb{E}\left[ \ell\paren{\mathbf{\hat{p}}_{trg}, \mathbf{p}_{trg} } \right]
\end{equation*}

The advantage of this method over the previous one is that the input to the inner network is of the order of magnitude of the number of phonemes, and not the number of features. This leads to more reasonable lengths of the inputs, but it relies more heavily on the model to learn to combine  feature representations correctly.

\section{Experiments}
\paragraph{Models}
We applied the described methods to two character-level models.\footnote{All our code is available at \url{https://github.com/OnlpLab/InflectionWithPhonology.git}} Both were modified to solve reinflection instead of inflection and to handle phonemic symbols and phonological features:

\begin{itemize}[leftmargin=1.5em]
    \item \emph{LSTM}: a standard LSTM Encoder-Decoder model with global attention\footnote{Not to be confused with the self-attention layer applied in the \emph{model manipulation} method.} as proposed in \citet{silfverberg-hulden-2018-encoder}.
    \item  \emph{Transduce}: An LSTM-based model by \citet{makarov-clematide-2018-neural} predicting edit actions between the source and the target. This model is more robust in low-resource settings.
\end{itemize}

\paragraph{Data} We experimented with eight languages: Swahili, Georgian, Albanian, Bulgarian, Latvian, Hungarian, Finnish and Turkish, in three part-of-speech types. All of these languages have shallow orthography, i.e., nearly one-to-one G2P and P2G mappings. We purposefully selected such languages to be able to disentangle the effects of convoluted orthographies from the potential benefits of phonetic decomposition to features, and to avoid the use of trainable G2P and P2G models that would inevitably propagate errors and serve as a confounding factor.
We compared the two proposed methods to the baseline where the models take letters as the source and target tokens.

We randomly sampled 10,000 reinflection samples from the UniMorph 3.0 repository \cite{mccarthy-etal-2020-unimorph} for train, validation and test sets, with 80\%-10\%-10\% splits ratios. The split was done such that the sets would have no overlapping lemmas, following \citet{goldman-etal-2022-un}. The models were trained separately for each language and POS.

\paragraph{Preprocessing}
Due to the orthographic shallowness of the selected languages we were able to implement for each language a rule-based component for G2P and P2G conversions. 

\paragraph{Evaluation} Two evaluation metrics are reported: exact match accuracy and averaged edit distance. For comparability, all predictions were measured at the grapheme level, by converting the predictions back to graphemes using the P2G component.\footnote{In case the conversion component could not find a matching phoneme to the sequence of features, it used an out-of-vocabulary token `\#'.}

\section{Results and Analysis}
Table \ref{tab:lemma-systems} shows the results of the two systems across the two methods, compared to the graphemic baseline, averaged over languages. The \emph{LSTM} model performs poorly, with 46 accuracy points at the baseline, and less than 30 points in the novel methods. The \emph{Transduce} model performed much better in general, with more than 80 points in all 3 settings. On average over the 15 language-POS combinations, training on our methods resulted in a slight drop of 2.8 points, which makes them comparable with the baseline. These results may imply that our methods fit better to stronger models, and that this setting and quantities may be considered as low-resource, at least without hallucination methods like that of \citet{anastasopoulos-neubig-2019-pushing}.

\begin{table}
\Huge
\centering
\setlength{\tabcolsep}{5pt}
\resizebox{\columnwidth}{!}{%
\begin{tabular}{c|ccc|c}
\toprule
\multirow{3}{*}{\textbf{Model}} & \multicolumn{3}{c}{\textbf{Method}} & \multicolumn{1}{|c}{\multirow{3}{*}{\textbf{Average}}} \\\cline{2-4} & \multirow{2}{*}{\textbf{Baseline}} & \textbf{Data} & \textbf{Model} & \\
& & \textbf{Manipulation} & \textbf{Manipulation} & \\
\midrule

LSTM & \textbf{46.5\%±0.8\%} & 26.4\%±0.5\% & 10.9\%±2.2\% & 27.9\%±0.5\% \\
Transduce & \textbf{83.6\%±0.2\%} & 80.3\%±0.2\% & 80.8\%±0.9\% & 81.6\%±0.2\% \\

\bottomrule
\end{tabular}%
}
 
\caption{Graphemic Accuracy of all systems, averaged on all language-POS datasets, and averaged over 3 seeds. Highest value per row is \textbf{bold}.}
\label{tab:lemma-systems}
\end{table}

Table \ref{tab:transduce-lemma} breaks down the results of the \emph{Transduce} model per language. In 7 out of 15 datasets, at least one of our methods outperformed the baseline. The difference varies from 0.9 up to 11.7 accuracy points. All in all, it seems that there is no connection between the relative success of the phonologically-aware methods and the abundance of morpho-phonological processes in a language. In Turkish, for instance, that has vowel harmony and additional phonological processes, the baseline performed much better, while in Swahili and Georgian (which barely exhibit such processes) there were clear improvements.

\begin{table}
\centering
\resizebox{\columnwidth}{!}{
\huge
\begin{tabular}{ll|ccc}
\toprule
\multirow{3}{*}{\textbf{Language}} & \multirow{3}{*}{\textbf{POS}} & \multicolumn{3}{c}{\textbf{Method}} \\\cline{3-5}
& & \multirow{2}{*}{\textbf{Baseline}} &  \textbf{Data} & \textbf{Model} \\
& & & \textbf{Manipulation} & \textbf{Manipulation} \\
\hline

Bulgarian & Adj &   \textbf{96.6\%±0.4\%} & 95.5\%±1.2\% & 95.7\%±2.4\% \\
Bulgarian & V &     \textbf{89.0\%±1.1\%} & 87.6\%±1.0\% & 88.0\%±1.5\% \\
Finnish & Adj &     \textbf{94.2\%±0.5\%} & 92.8\%±0.2\% & 92.8\%±0.1\% \\
Finnish & N &       82.3\%±0.8\% & \textbf{83.1\%±0.9\%} & 78.2\%±0.9\% \\
Finnish & V &       \textbf{88.1\%±2.1\%} & 79.8\%±2.8\% & 84.3\%±1.0\% \\
Hungarian & V &     \textbf{90.9\%±1.1\%} & 89.6\%±0.5\% & 89.7\%±0.8\% \\
Georgian & N &      90.2\%±0.5\% & \textbf{91.4\%±0.8\%} & 90.3\%±0.6\% \\
Georgian & V &      42.2\%±2.0\% & 28.4\%±1.5\% & \textbf{44.2\%±4.1\%} \\
Latvian & N &       88.4\%±0.8\% & \textbf{90.0\%±0.6\%} & 85.6\%±0.5\% \\
Latvian & V &       \textbf{76.5\%±0.9\%} & 70.9\%±0.9\% & 67.9\%±1.9\% \\
Albanian & V &      84.3\%±1.0\% & 79.6\%±1.4\% & \textbf{86.9\%±2.2\%} \\
Swahili & Adj &     66.7\%±2.9\% & \textbf{74.4\%±4.5\%} & 64.4\%±12.6\% \\
Swahili & V &       90.9\%±1.0\% & 87.0\%±2.1\% & \textbf{92.4\%±1.2\%} \\
Turkish & Adj &     \textbf{91.6\%±2.1\%} & 79.0\%±4.3\% & 76.8\%±2.3\% \\
Turkish & V &       \textbf{82.5\%±0.5\%} & 75.8\%±2.1\% & 74.9\%±0.9\% \\

\hline
\multicolumn{2}{c|}{\textbf{Average}} & \textbf{83.6\%±0.2\%} & 80.3\%±0.2\% & 80.8\%±0.9\% \\
\bottomrule
\end{tabular}
}

\caption{Graphemic Accuracy of the \emph{Transduce}, averaged over 3 seeds. Highest value per row is in \textbf{bold}.}
\label{tab:transduce-lemma}
\end{table}

To 
provide insights into the sufficiency of the data and the richness of the signal, we plot on figure~\ref{fig:lang-pos-LCs} (in appendix~\ref{sec:curves}) learning curves for the \emph{Transduce} model per language. We trained each model over an increasing number of train samples from 1,000 to 8,000 and evaluated them on the same test sets for each language.
The general trends show that the amount of data is indeed sufficient for the model and the signal is not richer, as in most cases the test accuracy with 8,000 samples is similar to the one with 3,000 samples.
Moreover, the graphs show that our methods have no clear advantage over the baseline even in as few as 1,000 training examples.


\section{Discussion and Conclusion}
\label{sec:discussion}
In this work we incorporated phonological information into morphological tasks. We proposed two methods: one that modifies the data, and one that also manipulates the model. We exemplified them on reinflection for two models 
and found out that, on average, our methods are comparable with the baseline and do not surpass it. 
We conclude that 
the embeddings obtained for the graphemic representations in such tasks may already encode the underlying phonological information in the data.

This conclusion is in line with the work of \citet{wiemerslage-etal-2018-phonological}, who similarly aimed, with no success, to use phonological data in morphological inflection.
Unlike our work, they used a weaker inflection model as a baseline for modification and they had a different method in constructing the 
phonologically-aware  embeddings. 
More crucially, they experimented with a \emph{form-split} setting, which means that there was significant overlap between the sampled lemmas in the train-test split.
Our results also corroborate the findings of \citet{silfverberg-etal-2018-sound}, who examined phoneme embeddings from various sources, including from a morphological inflection model, and showed that they implicitly encode phonological features, thus supporting our main conclusion.


\section*{Limitations}
One limitation of our work is the experimentation only with languages with shallow orthographies, i.e. relatively simple G2P and P2G mappings. The results might vary for deeper-orthographies languages.

Although we took extra care to verify our conversions are correct and complete, and designed the rules to be as comprehensive as possible, automatic rule-based processes in languages may not be 100\% perfect and some corner cases may introduce errors. These errors may propagate to affect the numerical results.
To mitigate this issue, when ambiguities in determining a target phoneme (or grapheme) in a given language occur, we purposefully select the values that occur more frequently in the UniMorph data of that particular language.

\section*{Acknowledgements}
This research is funded by a grant from
the European Research Council, ERC-StG grant
number 677352, and a grant by the Israeli Ministry of Science and Technology (MOST), grant number 3-17992, for which we are grateful.

\nocite{wu-etal-2021-applying}

\bibliography{anthology, custom}

\clearpage
\appendix
\section{Learning Curves}
\label{sec:curves}

\noindent\begin{minipage}{\textwidth}
    \centering
    \includegraphics[height=0.68\vsize, width=\hsize]{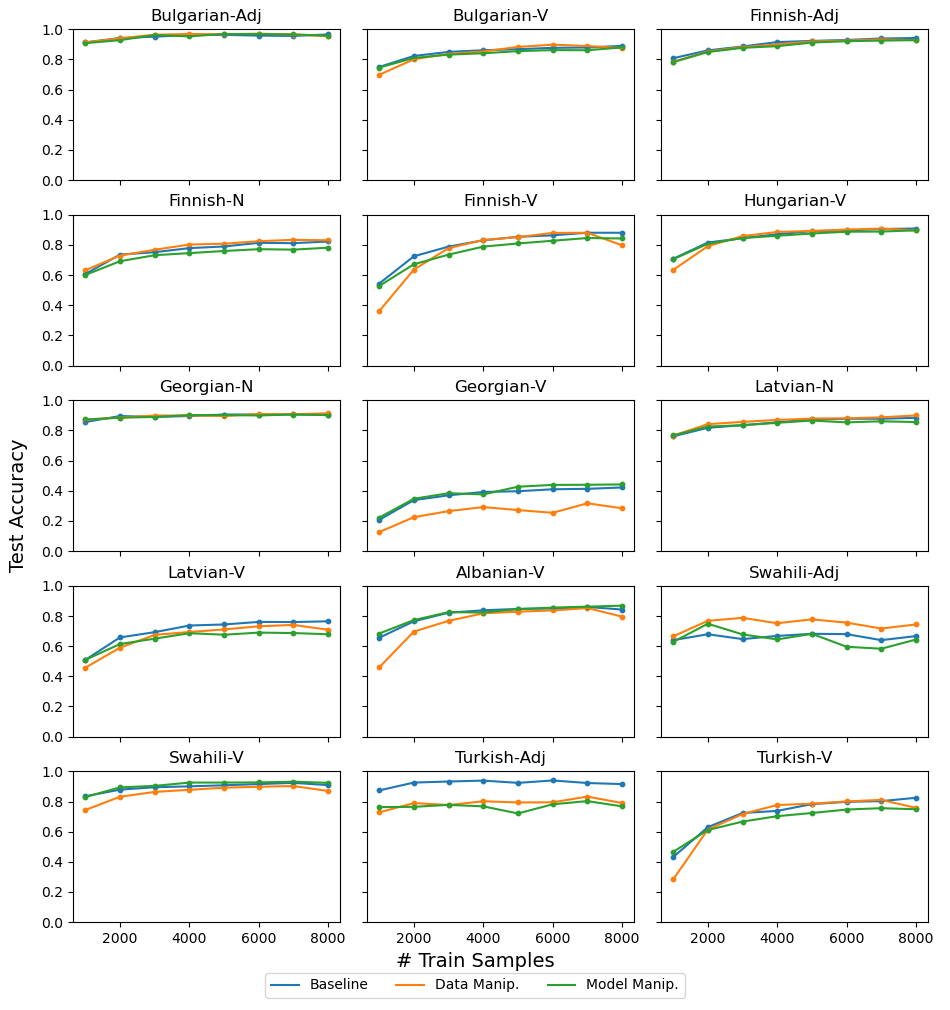}
    \captionof{figure}{Learning curves for accuracy over test sets for each language-POS dataset, as a function of the train set size.}
    \label{fig:lang-pos-LCs}
\end{minipage}

\end{document}